\documentclass{article}

\PassOptionsToPackage{numbers, compress}{natbib}

\usepackage[sglblindworkshop, preprint]{neurips_2026}

\usepackage[utf8]{inputenc} 
\usepackage[T1]{fontenc}    
\usepackage{hyperref}       
\usepackage{cleveref} 
\usepackage{url}            
\usepackage{booktabs}       
\usepackage{amsfonts}       
\usepackage{nicefrac}       
\usepackage{microtype}      
\usepackage{xcolor}         
\usepackage{multirow}
\usepackage{graphicx}
\usepackage{enumitem}

\title{Boundary-Aware Uncertainty Quantification \\ for Wildfire Spread Prediction}
\workshoptitle{Tackling Climate Change with Machine Learning}

\author{%
  Jonas V.~Funk \\
  Independent Research\\
 76199 Karlsruhe, Germany \\
  \texttt{jonasvilhofunk@gmail.com} \\
}

\begin{document}

\maketitle

\begin{abstract}
Reliable wildfire spread prediction is vital for risk-aware emergency planning, yet most deep learning models lack principled uncertainty quantification (UQ). Further, for boundary-sensitive cases like wildfire spread, evaluating models with global metrics alone is often insufficient. To shift the focus of UQ evaluation toward a more operationally relevant approach, the \textbf{Fire-Centered Evaluation Region (FCER)} framework is introduced as a spatially conditioned protocol to characterize UQ within critical fire zones. Using FCER, an Ensemble is compared against an distilled single-pass student model on the WildfireSpreadTS dataset. The student model demonstrates comparable calibration and complementary uncertainty ranking in boundary-relevant regimes. Code is available at \url{https://github.com/jonasvilhofunk/WildfireUQ-FCER}.
\end{abstract}

\section{Introduction}

Global warming is driving unprecedented fire behavior, placing reliable fire behavior prediction at the heart of effective disaster response~\cite{abram2021, haas2026}. Models should therefore be aware of their uncertainty during the most erratic propagation events, which are precisely the scenarios most critical for fire authorities.

While deep learning shows promise for wildfire spread prediction~\cite{andrianarivony2024, gerard2023}, existing approaches remain largely deterministic. Although \citet{chakravarty2025} identified Deep Ensembles as well‑calibrated for the WildfireSpreadTS~\cite{gerard2023} dataset, the linear scaling of inference cost with Ensemble members renders them impractical for latency‑sensitive real‑time fire response.

Recent work has shown that global, pixel‑wise uncertainty metrics obscure the spatial structure of errors~\cite{zeevi2025}, motivating evaluation protocols that explicitly account for fire geometry and boundary proximity. Efforts to characterize spatial uncertainty have reported buffer zones as narrow as 20--60~m~\cite{chakravarty2025} around firelines on WildfireSpreadTS. However, at the dataset's 375~m native resolution, such sub-pixel distances are difficult to interpret, motivating pixel-scale evaluation.

This paper evaluates two models on WildfireSpreadTS using the proposed \textbf{Fire-Centered Evaluation Region (FCER)} framework to enable fire-aware uncertainty quantification (UQ). The two compared models are a teacher Ensemble, as well as an uncertainty‑distilled student model. Contributions:

\begin{enumerate}[label=(\roman*)]
    \item Introduction of the \textbf{FCER framework} for evaluating UQ near fire boundaries through dilated ground truth fire masks, a radius sweep to capture spatial dependence, and an Average Surface Distance (ASD) based anchor for statistical comparison.
    \item Adaptation of Deep Uncertainty Distillation using Ensembles for Segmentation (DUDES)~\cite{landgraf2024} to train a student model for wildfire spread forecasting, enabling efficient single-pass uncertainty ranking while preserving calibration.
\end{enumerate}

\section{Methods}
This section outlines the experimental framework. Further details regarding metric definitions, implementation specifications, and ablation studies are provided in the Supplementary Material.

\textbf{Dataset:} WildfireSpreadTS is a 4‑year wildfire spread dataset covering 2018--2021 at 375~m resolution. The task is next‑day wildfire spread prediction from multi‑modal remote‑sensing data, yielding models that predict a probabilistic fire spread map for the following day. This work focuses on the $t=5$ setup, using the previous five days as input with only the vegetation channels, which prior work has found to be the most effective configuration~\cite{gerard2023, lahrichi2026}. Following \citet{lahrichi2026}, all metrics are computed within a central $128 \times 128$ pixel crop.

\textbf{Model:} A U‑Net with Temporal Attention Encoder (UTAE)~\cite{garnot2021} is used as the main backbone, designed for spatiotemporal segmentation and shown to perform best on WildfireSpreadTS~\cite{gerard2023, lahrichi2026}.

\textbf{Ensemble:} Serves as a baseline and consists of $n$ independently trained UTAE models (pretrained checkpoints of \citet{lahrichi2026}), whose predictions are averaged at test time. It is worth noting that inference cost scales linearly with the number of ensemble members, since each model requires a separate forward pass. A pixel-wise uncertainty map is computed from disagreement across member predictions, yielding higher uncertainty in regions where the models are less consistent. The ensemble size is chosen as $n=3$, based on the leave-one-year-out fold construction: when one year is held out for testing, the remaining years yield exactly three non-overlapping train/validation splits.

\textbf{DUDES:} The uncertainty prediction from the Ensemble is distilled into a single DUDES student model. The students backbone (pretrained checkpoints of \citet{lahrichi2026}) is frozen and a lightweight uncertainty head (a $1 \times 1$ convolution followed by sigmoid) is appended to the decoder's final feature map. This isolates uncertainty discrimination from the feature extraction already encoded in the backbone, and departs from the original DUDES formulation~\cite{landgraf2024}, which retrains the full network. Training minimizes Root Mean Squared Log Error against the teacher's normalized uncertainty prediction.

\textbf{Metrics:} \textit{Segmentation} quality is reported using Average Precision (AP) for detection performance and Average Surface Distance (ASD) for boundary alignment. ASD quantifies the symmetric distance between predicted and ground-truth fire boundaries, with lower values indicating better spatial alignment. \textit{Calibration} is assessed via the Brier score and Negative Log-Likelihood (NLL). \emph{UQ~ranking} is evaluated using Area Under the Receiver Operating Characteristic Curve (AUROC) and Area Under the Precision-Recall Curve (AUPRC), treating misclassified pixels as positives and correctly classified pixels as negatives. Higher uncertainty should rank error pixels above correctly predicted ones. For both models, this ranking is computed against the same error map derived from the middle‑AP ensemble member’s predictions, ensuring fair comparison. The AUPRC random baseline equals the prevalence of error pixels.

\textbf{FCER framework:} To evaluate models UQ spatially, a three-part protocol is employed:

(i) \emph{Fire-Centered Evaluation Region:} A spatial mask is defined by dilating the ground truth fire mask $\Omega_{\mathrm{GT}}$ by a radius $r_d$. Formally, the FCER is defined as 
$\Omega_{\mathrm{eval}} = \Omega_{\mathrm{GT}} \oplus B_{r_d}$, where $B_{r_d}$ is a disk-shaped structuring element. This restricts evaluation to a neighborhood surrounding the fire boundary, effectively mitigating bias from large areas of background.

(ii) \emph{FCER sweep:} The dilation radius $r_d$ is systematically varied, revealing how UQ evolves as the evaluation neighborhood expands outward from the fire edge, exposing spatial structure that fixed-region metrics often obscure. In this work, \emph{boundary‑aware UQ} is operationalized as the dependence of uncertainty ranking (AUROC/AUPRC) on the dilation radius around the GT fire mask, as measured by the FCER sweep.

(iii) \emph{ASD anchor:} To provide a robust summary metric for model comparison, UQ is reported at a fixed anchor point: FCER with $r_d = \mathrm{ASD}$, which captures the typical scale of the models’ boundary prediction error. Statistical comparison between models is performed at this anchor using a paired per‑fire‑image Wilcoxon signed‑rank test \cite{wilcoxon1945}. Effect size is reported as the rank‑biserial correlation $r \in [-1, 1]$. Larger positive values indicate stronger evidence in favor of DUDES.

\section{Results}
The Ensemble teacher and the DUDES student are evaluated using the FCER framework to assess their boundary‑aware UQ.

\textbf{Segmentation:} Both models achieve similar AP and ASD, with mean ASD ($\approx 1.4~\mathrm{km}$) serving as the anchor for boundary-scale FCER evaluation (\Cref{tab:1}).

\textbf{Calibration:} At the ASD anchor $r_d = \mathrm{ASD}$, the Brier score and NLL are comparable, indicating that DUDES retains the Ensemble’s calibration~(\Cref{tab:1}). This pattern remains consistent across the FCER sweep, as shown in the Supplementary Material.

\begin{figure}[ht]
  \centering
  \includegraphics[width=\linewidth]{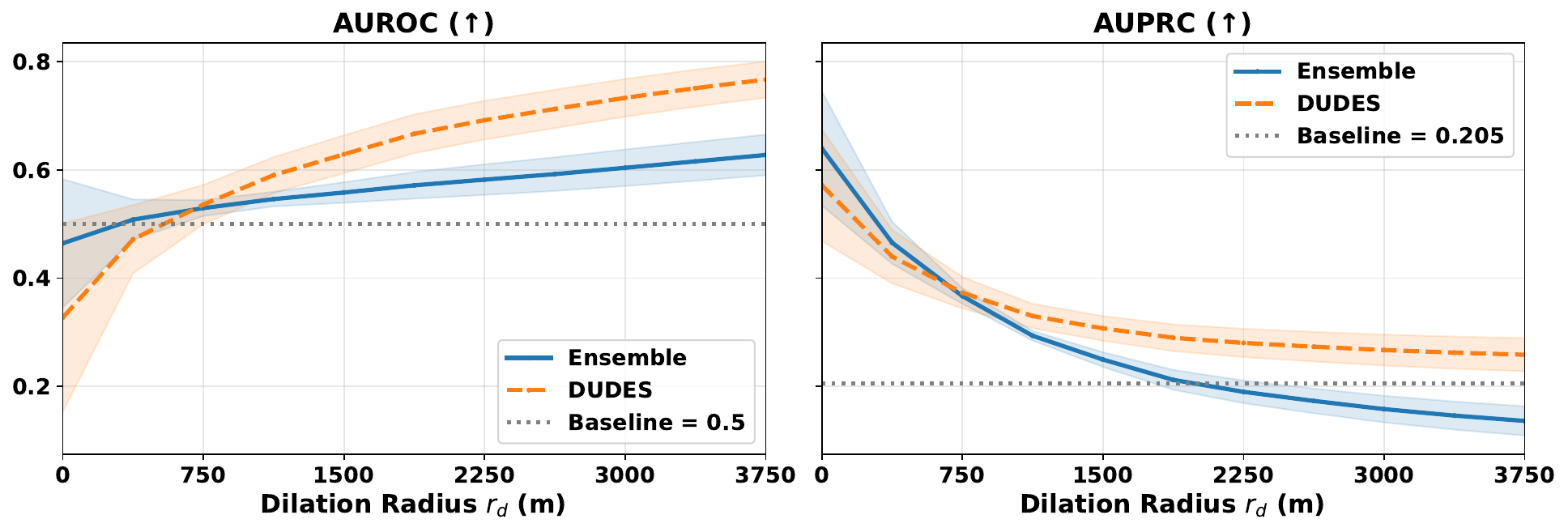}
\caption{FCER sweep on WildfireSpreadTS averaged over 2018--2021 for Ensemble and DUDES with UTAE (t=5) backbones. AUROC and AUPRC are computed within the FCER while varying \(r_d\).}
  \label{fig:1}  
\end{figure}

\textbf{UQ ranking:} At the ASD anchor, DUDES outperforms the Ensemble on both AUROC and AUPRC across all years (\Cref{tab:1}), though the Ensemble leads at very narrow $r_D < 750~\rm{m}$ (\Cref{fig:1}). Relative to the random baselines (AUROC: 0.5; AUPRC: 0.205), DUDES achieves mean AUROC of 0.629 (+26\%) and AUPRC of 0.307 (+50\%), compared to the Ensemble's 0.558 (+12\%) and 0.249 (+23\%). DUDES thus operates further above chance, indicating more effective uncertainty ranking. Pooled Wilcoxon signed-rank tests confirm that this advantage is statistically significant across all years for both AUROC (\(r = +0.78\), \(p < 0.001\)) and AUPRC (\(r = +0.69\), \(p < 0.001\)).

\begingroup
\setlength{\tabcolsep}{3pt}
\begin{table}[ht]
\caption{Per-year results on WildfireSpreadTS. Calibration metrics (Brier score, NLL) and UQ ranking metrics (AUROC, AUPRC) are computed within an FCER obtained by dilating the ground truth fire mask by $1\times$ ASD.}
  \centering
  \begin{tabular}{ll cc cc cc}
    \toprule
    & & \multicolumn{2}{c}{Segmentation} & \multicolumn{2}{c}{Calibration @$r_d= \mathrm{ASD}$} & \multicolumn{2}{c}{UQ ranking @$r_d= \mathrm{ASD}$} \\
    \cmidrule(r){3-4} \cmidrule(lr){5-6} \cmidrule(l){7-8}
    Year & Method & AP $\uparrow$ & ASD [km] $\downarrow$ & Brier $\downarrow$ & NLL $\downarrow$ & AUROC $\uparrow$ & AUPRC $\uparrow$ \\
    \midrule
    \multirow{2}{*}{2018}
      & Ensemble & 0.53 & 1.21 & 0.159 & 0.505 & 0.562 & 0.251 \\
      & DUDES    & 0.51 & 1.22 & 0.160 & 0.510 & \textbf{0.603} & \textbf{0.281} \\
    \addlinespace
    \multirow{2}{*}{2019}
      & Ensemble & 0.37 & 1.13 & 0.163 & 0.517 & 0.527 & 0.233 \\
      & DUDES    & 0.35 & 1.22 & 0.169 & 0.541 & \textbf{0.619} & \textbf{0.318} \\
    \addlinespace
    \multirow{2}{*}{2020}
      & Ensemble & 0.51 & 1.52 & 0.173 & 0.549 & 0.568 & 0.270 \\
      & DUDES    & 0.50 & 1.71 & 0.172 & 0.548 & \textbf{0.605} & \textbf{0.291} \\
    \addlinespace
    \multirow{2}{*}{2021}
      & Ensemble & 0.60 & 1.68 & 0.150 & 0.476 & 0.577 & 0.243 \\
      & DUDES    & 0.58 & 1.51 & 0.151 & 0.481 & \textbf{0.689} & \textbf{0.339} \\
    \midrule
    \multirow{2}{*}{Mean}
      & Ensemble & $0.50_{\pm0.08}$ & $1.39_{\pm0.22}$ & $0.161_{\pm0.008}$ & $0.512_{\pm0.026}$ & $0.558_{\pm0.019}$ & $0.249_{\pm0.014}$ \\
      & DUDES    & $0.49_{\pm0.09}$ & $1.41_{\pm0.21}$ & $0.163_{\pm0.008}$ & $0.520_{\pm0.027}$ & $\mathbf{0.629_{\pm0.035}}$ & $\mathbf{0.307_{\pm0.023}}$ \\
    \bottomrule
  \end{tabular}
  \label{tab:1}
\end{table}
\endgroup

\textbf{FCER sweep:} \Cref{fig:1} shows DUDES outperforming the Ensemble uncertainty ranking beyond 750~m, with the gap widening at larger radii. This reveals complementary strengths: the Ensemble excels very close to the boundary, while DUDES generalizes better over error‑relevant neighborhoods. Similar trends hold for different backbone variants, shown in Supplementary Material.

\begin{figure}[ht]
  \centering
  \includegraphics[width=\linewidth]{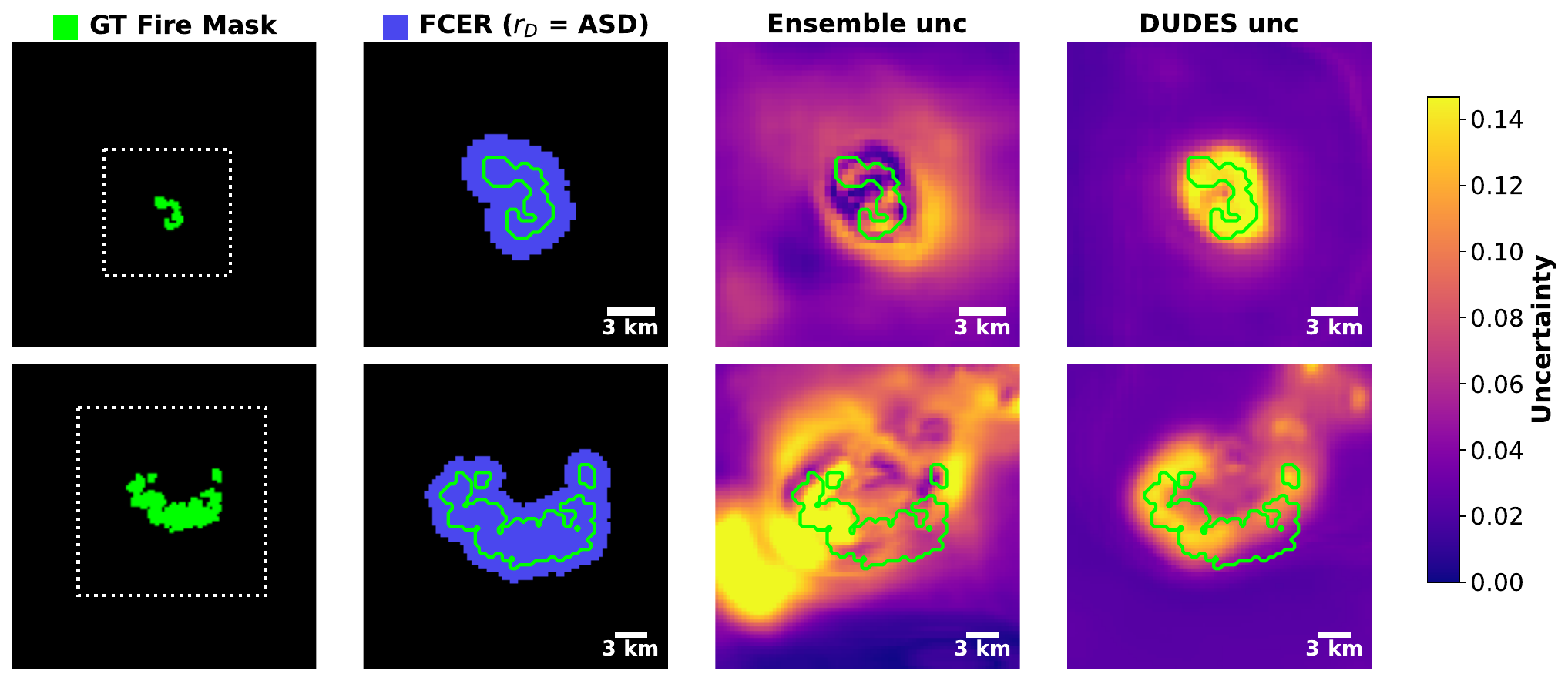}
\caption{Representative uncertainty (unc) maps for a small fire (top) and a large fire (bottom) from 2021.  First column: full scene; Rest: cropped region. Cases are chosen as representative samples with a $\Delta$AUROC closest to the annual mean at the $r_d=\mathrm{ASD}$ anchor.}
  \label{fig:2}  
\end{figure}

\textbf{Qualitative visualization:} \Cref{fig:2} shows two wildfire events 
from 2021. DUDES uncertainty appears more compact and boundary-focused than 
that of the Ensemble, which is more diffuse and noisy, particularly in 
regions farther from the fire boundary.

\section{Discussion}
These results highlight the ability of the FCER framework to expose spatially dependent UQ behaviors that remain hidden under global metrics.

\textbf{Discussion:} The FCER sweep clearly shows that UQ behavior depends strongly on distance from the fire edge. The Ensemble is stronger very close to the boundary, likely because its member disagreement better captures fine-grained boundary ambiguity. DUDES appears stronger around the ASD anchor and beyond, possibly because distillation emphasizes broader neighborhood patterns over fine-grained ensemble disagreement. Furthermore, DUDES matches the Ensemble in segmentation and calibration while requiring only a single forward pass, yielding lower inference cost. As the evaluation region expands, both models improve because the background becomes easier to classify than the boundary region, reducing overall uncertainty and error. Within the FCER framework, DUDES provides a practical single-pass alternative that preserves calibration while improving uncertainty ranking in boundary-relevant regions.

\textbf{Limitations:} The evaluation uses $n=3$ Ensemble members and a $128 \times 128$ center-region crop, tying the results to that specific spatial window and ensemble size. The backbone is frozen by design, how end-to-end retraining would affect DUDES boundary-aware UQ remains open. While the observed ASD was similar across models and the mean value was used as the anchor, the sensitivity of FCER to highly divergent model-specific ASD values has not yet been examined. Furthermore, the analysis is limited to WildfireSpreadTS; broader comparisons, including UQ methods such as diffusion-based approaches, would provide a more complete picture.

\section{Climate Change Impact}
Projections predict increases in global fire size of 23--36\% and mean global fire intensity of 2--5\% even under ambitious climate mitigation \cite{haas2026}. This makes uncertainty estimates near the fireline operationally critical for time-sensitive response decisions. The FCER framework evaluates models in this boundary region directly, and DUDES offers a single-pass alternative to ensemble-scale inference in resource-constrained systems.

\begin{ack}
This work received no external funding. The author declares no competing interests.
\end{ack}

\bibliographystyle{unsrtnat}  
\bibliography{references}

\newpage

\appendix

\section{Supplementary Material}
\label{sec:SupplementaryMaterial}
This appendix is organized as follows. \Cref{sec:S1} defines all metrics used in the paper. \Cref{sec:S2} provides additional FCER sweep plots omitted from the main paper for space, including per-year AUROC and AUPRC curves as well as mean calibration behavior (Brier score and NLL) across radii for the UTAE ($t=5$) backbone. \Cref{sec:S3} reports ablation experiments on U-Net ($t=1$) and U-Net ($t=5$) variants to assess whether the DUDES distillation and FCER protocol generalize across backbone architectures. \Cref{sec:S4} provides implementation details.

\subsection{Metric Definitions}
\label{sec:S1}

Let \(y_i \in \{0,1\}\) denote the ground-truth label for pixel \(i\), \(\hat{p}_i \in [0,1]\) the predicted fire probability, and \(N\) the number of evaluated pixels. For segmentation metrics requiring binary masks, probabilistic predictions are thresholded at \(0.5\) to obtain a predicted mask \(\hat{Y}\), which is compared against the ground-truth mask \(Y\).

\subsubsection{Segmentation metrics}

\textbf{Precision:}
Precision is the fraction of predicted positive pixels that are truly positive:
\[
\mathrm{Precision} = \frac{\mathrm{TP}}{\mathrm{TP}+\mathrm{FP}}.
\]

\textbf{Recall:}
Recall is the fraction of true positive pixels that are correctly identified:
\[
\mathrm{Recall} = \frac{\mathrm{TP}}{\mathrm{TP}+\mathrm{FN}}.
\]

\textbf{Average Precision (AP):}
Average Precision summarizes the precision--recall curve obtained by thresholding the predicted fire probabilities over all possible decision thresholds. It measures how well the model ranks fire pixels above non-fire pixels. \emph{Higher values indicate better detection performance.}

\textbf{Average Surface Distance (ASD):}
Average Surface Distance, also referred to as ASSD in parts of the segmentation literature, measures the symmetric boundary discrepancy between the predicted binary fire mask \(\hat{Y}\) and the ground-truth mask \(Y\). Let \(\partial \hat{Y}\) and \(\partial Y\) denote their boundary pixel sets, and let \(d(a,B)\) be the Euclidean distance from a boundary point \(a\) to the closest point in boundary set \(B\). The ASD is defined as
\[
\mathrm{ASD}(\hat{Y},Y)
=
\frac{1}{2}
\left(
\frac{1}{|\partial \hat{Y}|}\sum_{a \in \partial \hat{Y}} d(a,\partial Y)
+
\frac{1}{|\partial Y|}\sum_{b \in \partial Y} d(b,\partial \hat{Y})
\right).
\]
Distances are reported in meters after conversion from pixel units using the dataset resolution. \emph{Lower values indicate better boundary alignment.}

\subsubsection{Calibration metrics}

\textbf{Brier score:}
The Brier score measures the mean squared error between predicted probabilities and binary ground-truth labels:
\[
\mathrm{Brier}
=
\frac{1}{N}\sum_{i=1}^{N}(\hat{p}_i-y_i)^2.
\]
\emph{Lower values indicate better probabilistic predictions.}

\textbf{Negative Log-Likelihood (NLL):}
Negative Log-Likelihood measures the log-loss of the predicted Bernoulli probabilities. For numerical stability, predicted probabilities \(\hat{p}_i\) are clipped to \([\epsilon, 1-\epsilon]\), yielding \(\tilde{p}_i = \mathrm{clip}(\hat{p}_i, \epsilon, 1-\epsilon)\). The NLL is then defined as
\[
\mathrm{NLL}
=
-\frac{1}{N}\sum_{i=1}^{N}
\left[
y_i \log(\tilde{p}_i)
+
(1-y_i)\log(1-\tilde{p}_i)
\right].
\]
\emph{Lower values indicate better probabilistic predictions.}

\subsubsection{Uncertainty ranking metrics}

\textbf{Area Under the Receiver Operating Characteristic Curve (AUROC):}
AUROC measures how well uncertainty separates misclassified pixels from correctly classified pixels. Let \(u_i\) denote the uncertainty assigned to pixel \(i\), and let \(e_i \in \{0,1\}\) indicate whether that pixel is misclassified under the reference prediction. By sweeping a threshold over \(u_i\), a Receiver Operating Characteristic curve is formed from the true positive rate and false positive rate. AUROC is the area under this curve. \emph{Higher values indicate better ranking of error pixels above correct pixels.}

\textbf{Area Under the Precision-Recall Curve (AUPRC):}
AUPRC evaluates the same ranking task as AUROC, but summarizes the precision--recall curve instead. Here, the positive class is the set of misclassified pixels and the negative class is the set of correctly classified pixels. \emph{Higher values indicate that highly uncertain pixels are more likely to correspond to true errors.} The random baseline for AUPRC equals the prevalence of error pixels in the evaluated region.

\textbf{Statistical significance testing:}
To compare DUDES against the Ensemble, performance differences at the ASD anchor are assessed using a one-sided Wilcoxon signed-rank test~\cite{wilcoxon1945} applied to paired per-fire AUROC and AUPRC values, testing whether DUDES achieves higher scores than the Ensemble. The test is applied to the pooled set of fires across years. Tied zero-differences are discarded, following the default Wilcoxon signed-rank procedure. Effect size is reported as the rank-biserial correlation,
\[
r = \frac{W^{+}-W^{-}}{W^{+}+W^{-}},
\]
where \(W^{+}\) and \(W^{-}\) are the sums of ranks for positive and negative paired differences, respectively. \emph{Larger positive values indicate stronger evidence in favor of DUDES.}

\subsection{Additional FCER Sweep plots}
\label{sec:S2}

This section provides additional FCER sweep plots omitted from the main paper for space reasons. The figures complement the aggregate results by showing year-specific uncertainty ranking behavior as well as calibration trends.

\subsubsection{Per-year uncertainty ranking (AUROC and AUPRC, UTAE ($t=5$))}
\begin{figure}[ht]
  \centering
  \includegraphics[width=\linewidth]{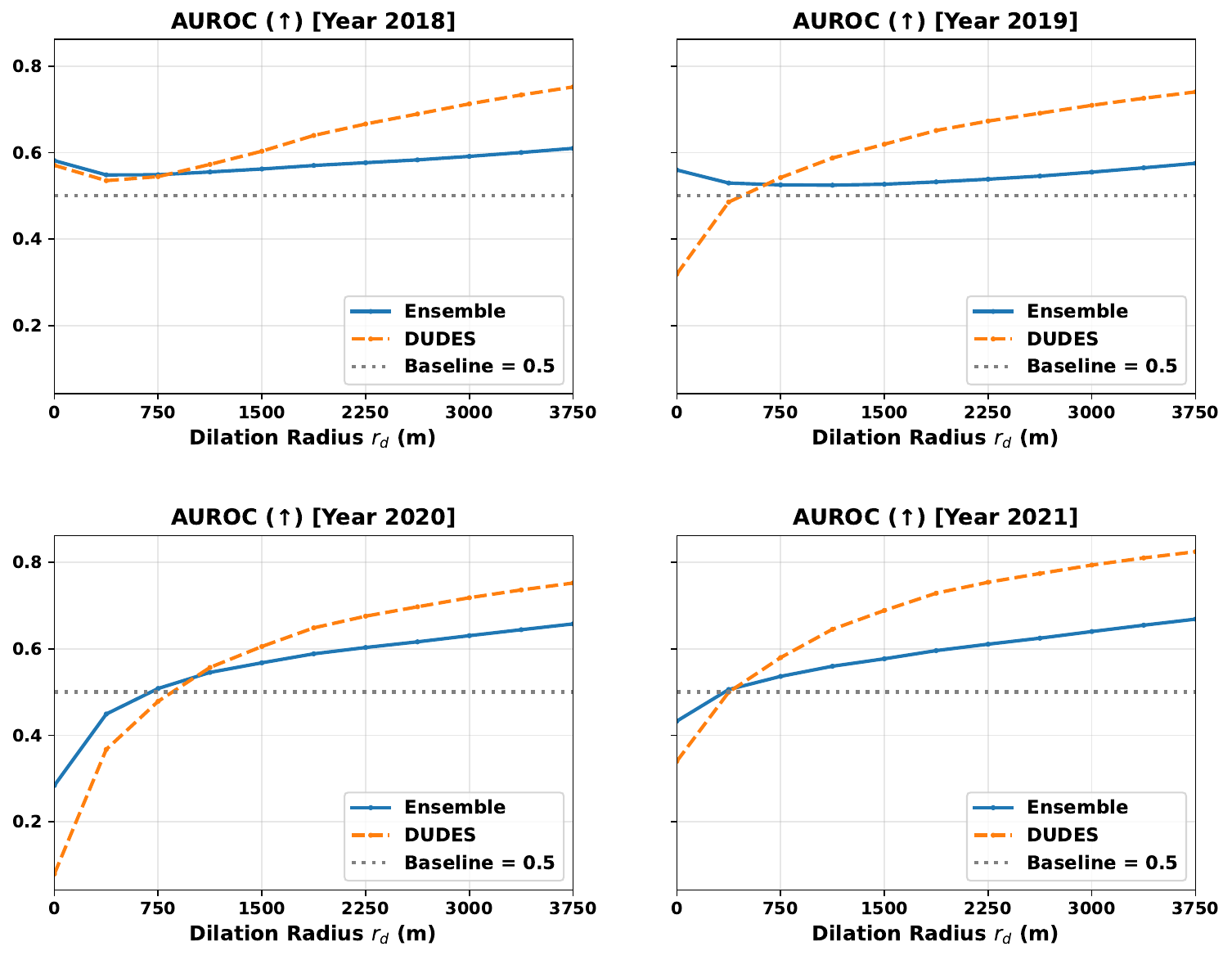}
    \caption{FCER sweep on WildfireSpreadTS for Ensemble and DUDES with UTAE ($t=5$) backbones, shown separately for each year from 2018 to 2021. AUROC is computed within the FCER while varying \(r_d\).}
  \label{fig:S1}  
\end{figure}

\begin{figure}[ht]
  \centering
  \includegraphics[width=\linewidth]{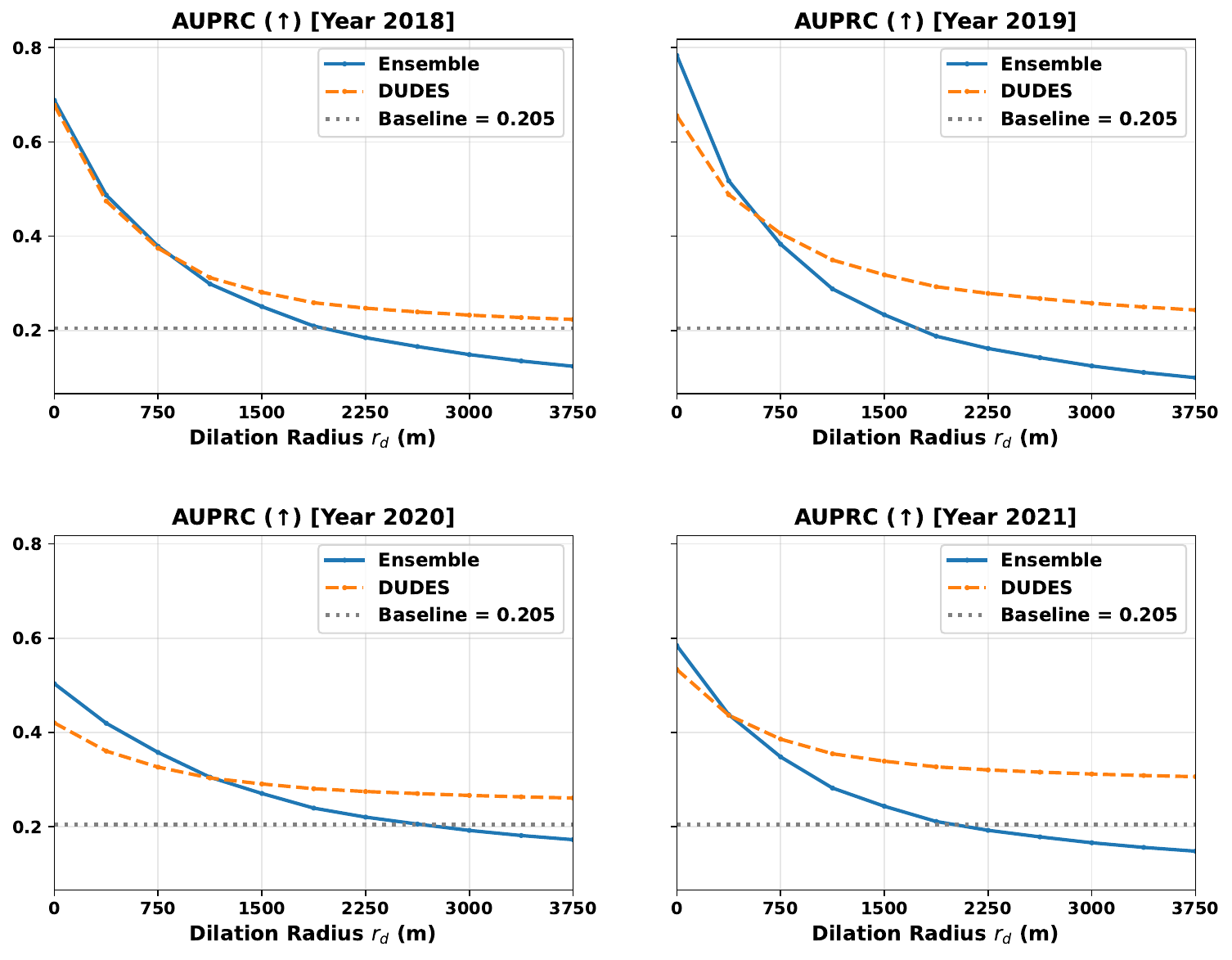}
    \caption{FCER sweep on WildfireSpreadTS for Ensemble and DUDES with UTAE ($t=5$) backbones, shown separately for each year from 2018 to 2021. AUPRC is computed within the FCER while varying \(r_d\).}
  \label{fig:S2}  
\end{figure}

\Cref{fig:S1} and \Cref{fig:S2} report the FCER sweep separately for each test year.

\subsubsection{Mean Calibration behavior over radius (Brier score and NLL, UTAE ($t=5$)}
\begin{figure}[ht]
  \centering
  \includegraphics[width=\linewidth]{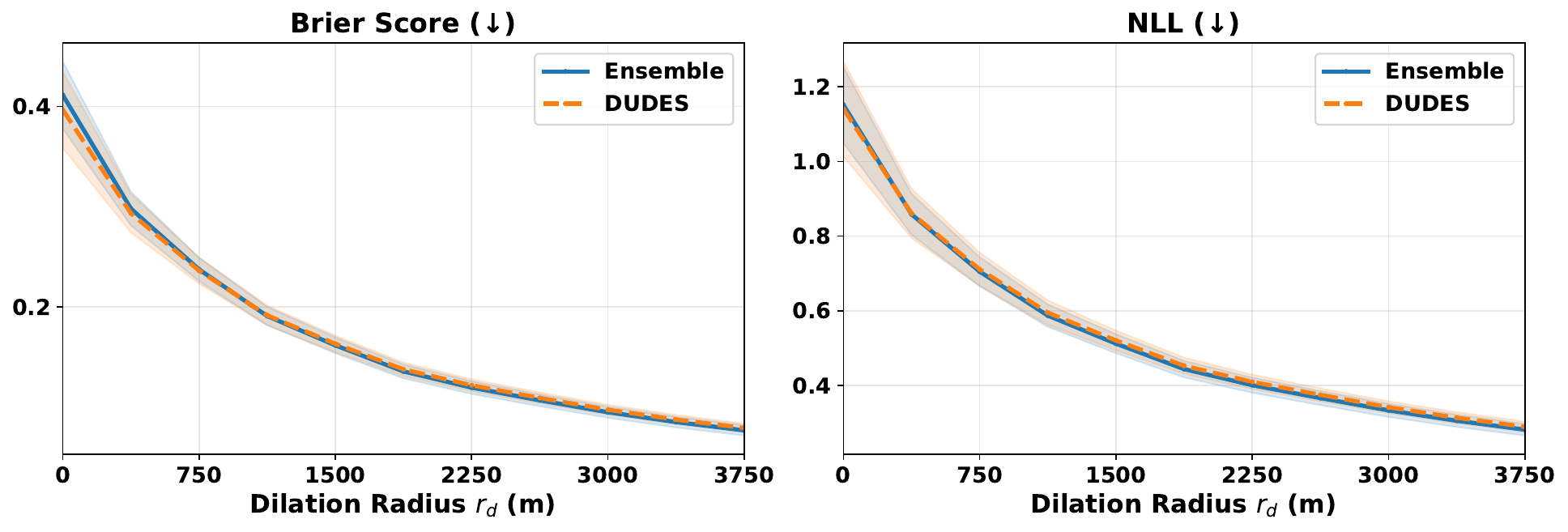}
  \caption{FCER sweep on WildfireSpreadTS for Ensemble and DUDES with UTAE ($t=5$) backbones, averaged over 2018 to 2021. Brier score and NLL are computed within the FCER while varying \(r_d\).}
  \label{fig:S3}  
\end{figure}

\Cref{fig:S3} shows the mean calibration behavior across radii. In contrast to the stronger differences observed for uncertainty ranking, calibration remains similar between Ensemble and DUDES throughout the FCER sweep.

\subsection{Model-variant sensitivity}
\label{sec:S3}
To assess whether the FCER trends observed for UTAE generalize beyond the main backbone, additional experiments are reported for U-Net variants with \(t=1\) and \(t=5\). As in the main experiments, Ensemble teachers are constructed from pretrained checkpoints released by \citet{lahrichi2026}, and DUDES is evaluated against the corresponding Ensemble baseline. These results serve as a robustness check on the FCER protocol rather than as a full backbone benchmark.

\subsubsection{U-Net \(t=1\)}
\begin{figure}[ht]
  \centering
  \includegraphics[width=\linewidth]{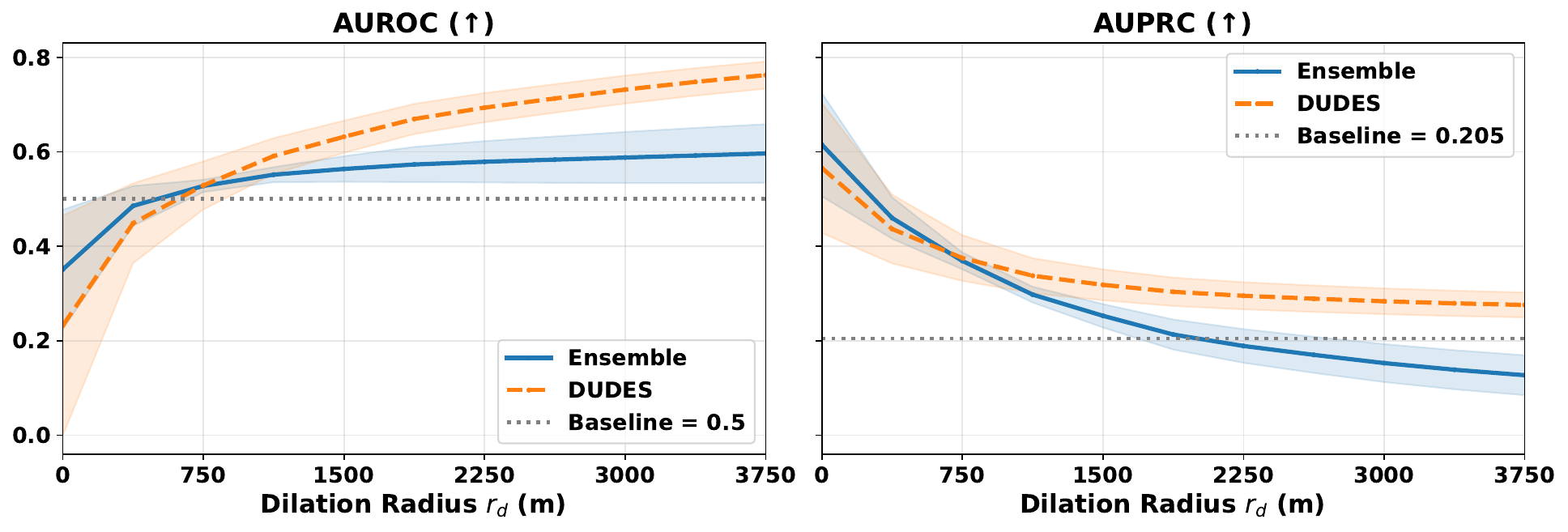}
  \caption{FCER sweep on WildfireSpreadTS averaged over 2018--2021 for Ensemble and DUDES with U-Net (\(t=1\)) backbones. AUROC and AUPRC are computed within the FCER while varying \(r_d\).}
  \label{fig:S4}  
\end{figure}

Shown in \Cref{fig:S4}, the FCER sweep for the U-Net (\(t=1\)) variant is qualitatively consistent with the main UTAE results.

\subsubsection{U-Net \(t=5\)}
\begin{figure}[ht]
  \centering
  \includegraphics[width=\linewidth]{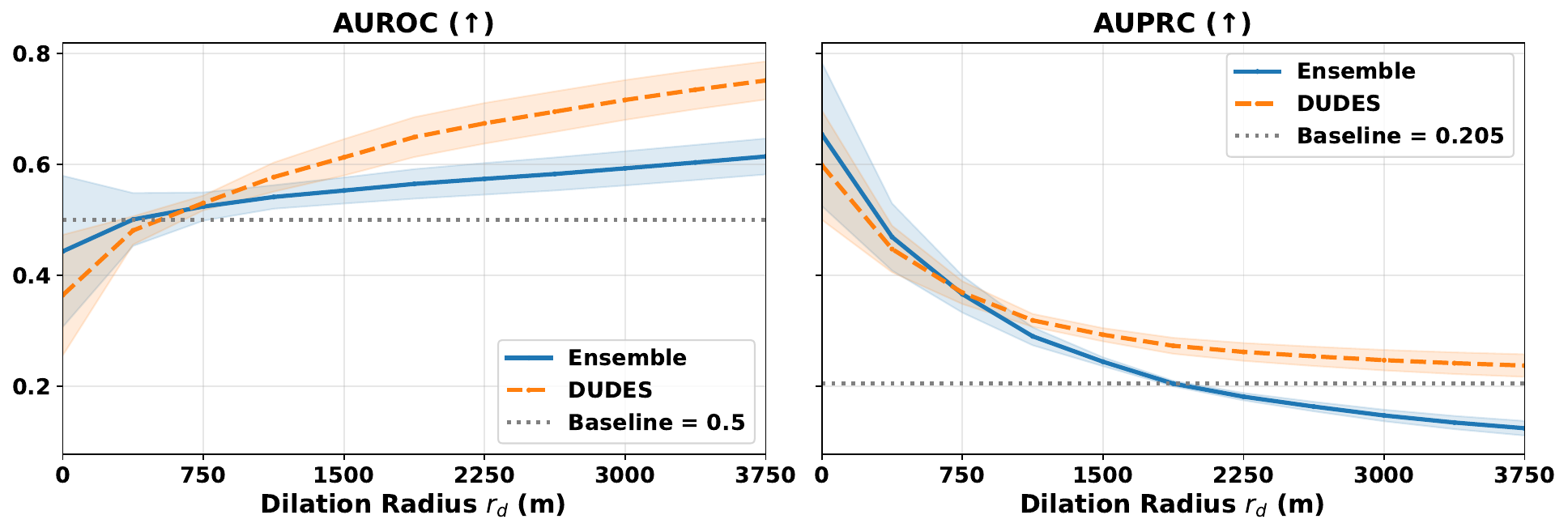}
  \caption{FCER sweep on WildfireSpreadTS averaged over 2018--2021 for Ensemble and DUDES with U-Net (\(t=5\)) backbones. AUROC and AUPRC are computed within the FCER while varying \(r_d\).}
  \label{fig:S5}  
\end{figure}

Shown in \Cref{fig:S5}, the FCER sweep for the U-Net (\(t=5\)) variant is likewise qualitatively consistent with the main UTAE results.

\subsection{Implementation Details}
\label{sec:S4}

The main experiment uses the pretrained UTAE (\(t=5\)) checkpoints released by \citet{lahrichi2026} on WildfireSpreadTS.

\textbf{Fold construction:}
For each test year, the Ensemble teacher is formed from the three fair folds obtained from the remaining three years under all valid train/validation assignments. For example, when 2018 is the test year, the three folds correspond to training on \{2019, 2020\} with validation on 2021, training on \{2019, 2021\} with validation on 2020, and training on \{2020, 2021\} with validation on 2019. The DUDES student is attached to the backbone with median AP among these three folds.

\textbf{Teacher uncertainty and cached training:}
Teacher uncertainty is computed as the normalized per-pixel sample standard deviation across the three Ensemble probability maps, scaled by the theoretical maximum for $n=3$. Only the lightweight uncertainty head is trained, while the backbone remains frozen. The head is attached non-intrusively via a PyTorch forward hook registered on the decoder's final feature map, requiring no modification to the frozen backbone's computational graph. To reduce computation, decoder features and teacher uncertainty targets are cached once for the training and validation sets, so optimization is performed only on the cached tensors.

\textbf{Training objective:}
Training minimizes the Root Mean Squared Logarithmic Error (RMSLE) between the student-predicted uncertainty and the teacher uncertainty target:
\[
\mathcal{L}_{\mathrm{RMSLE}}
=
\sqrt{
\frac{1}{N}\sum_{i=1}^{N}
\left(
\log\!\bigl(t_i(x)+1\bigr)-\log\!\bigl(s_i(x)+1\bigr)
\right)^2
}.
\]
For image \(x\), \(N\) is the number of pixels, \(t_i(x)\) is the teacher uncertainty target at pixel \(i\), and \(s_i(x)\) is the corresponding student prediction.

\textbf{Optimization and model selection:}
Optimization uses stochastic gradient descent (SGD) with momentum \(0.9\), weight decay \(5 \times 10^{-4}\), an initial learning rate of \(10^{-3}\), batch size 4, and a polynomial learning-rate decay schedule, with early stopping after 20 epochs without improvement. The retained checkpoint is the one achieving the highest validation AUROC within the FCER at the Ensemble ASD anchor (\(r_d=\mathrm{ASD}_{\mathrm{val}}=4\) px), fixed prior to test evaluation.

Full training and evaluation code is available at \url{https://github.com/jonasvilhofunk/WildfireUQ-FCER}.



\end{document}